\RequirePackage{fix-cm}
\documentclass[twocolumn]{svjour3}          
\smartqed  

\usepackage[misc]{ifsym}
\usepackage{epsfig}
\usepackage{graphicx}
\usepackage[tbtags]{amsmath}
\usepackage{amssymb}
\usepackage{float}
\usepackage{subfigure}
\usepackage{threeparttable}
\usepackage{framed}
\usepackage{overpic}
\usepackage{multirow}
\usepackage{array}
\usepackage{enumitem}
\usepackage{bm}
\usepackage[ruled, linesnumbered]{algorithm2e}
\usepackage[pagebackref=true,breaklinks=true,letterpaper=true,colorlinks,bookmarks=false,citecolor=blue]{hyperref}
\usepackage{natbib}
\usepackage{booktabs}
\usepackage{color}

\usepackage{etoolbox}
\newcommand*{\affaddr}[1]{#1} 
\newcommand*{\affmark}[1][*]{\textsuperscript{#1}}

\begin{document}

\title{Separating Content from Style Using Adversarial Learning \\ for 
Recognizing  Text
in the Wild}


\author{
Canjie Luo \and Qingxiang Lin \and Yuliang Liu \and Lianwen Jin \and Chunhua Shen
}

%
%

\institute{
	\Letter~Lianwen Jin\affmark[1,4] \at \email{eelwjin@scut.edu.cn}
	\and 
	   Chunhua Shen\affmark[2,3]      \at \email{chunhua.shen@adelaide.edu.au}
	\and 
	   Canjie Luo\affmark[1]          \at \email{canjie.luo@gmail.com}
	\and
	   Qingxiang Lin\affmark[1]       \at \email{qingxiang.lin@foxmail.com}
	\and
	   Yuliang Liu\affmark[1,2]       \at \email{liu.yuliang@mail.scut.edu.cn} 
	\and
	\affaddr{\affmark[1]South China University of Technology}\\
	\affaddr{\affmark[2]The University of Adelaide}
	\affaddr{\affmark[3]Monash University}\\
	\affaddr{\affmark[4]SCUT-Zhuhai Institute of Modern Industrial Innovation}
}

\date{Received: date / Accepted: date}

\maketitle

\begin{abstract}
Scene text  recognition is an important task in computer vision.  Despite tremendous progress achieved in the past few years, issues such as  varying font styles, arbitrary shapes and complex backgrounds etc.\ have made the problem very challenging. In this work, we propose to improve text recognition from a new perspective by separating the text content from complex backgrounds, thus making the recognition considerably easier and significantly improving recognition accuracy. 

To this end, we exploit the generative adversarial networks (GANs) for removing backgrounds while retaining the text content. As vanilla GANs are not sufficiently robust to generate sequence-like characters in natural images, we propose an adversarial learning framework for the generation and recognition of multiple characters in an image. The proposed framework consists of an attention-based recognizer and a generative adversarial architecture. Furthermore, to tackle the issue of lacking paired training samples, we design an interactive joint training scheme, which shares attention masks from the recognizer to the discriminator, and enables the discriminator to extract the features of each character for further adversarial training. Benefiting from the character-level adversarial training, our framework requires only unpaired simple data for style supervision. Each target style sample containing only one randomly chosen character can be simply synthesized online during the training. This is significant as the training does not require costly paired samples or character-level annotations. Thus, only the input images and corresponding text labels are needed. In addition to the style normalization of the backgrounds, we refine character patterns to ease the recognition task. A feedback mechanism is proposed to bridge the gap between the discriminator and the recognizer. Therefore, the discriminator can guide the generator according to the confusion of the recognizer, so that the generated patterns are clearer for recognition. Experiments on various benchmarks, including both regular and irregular text, demonstrate that our method significantly reduces the difficulty of recognition. Our framework can be integrated into recent recognition methods to achieve new state-of-the-art recognition accuracy.

\keywords{
    Text recognition 
    \and
    Attention mechanism 
    \and
    Generative adversarial network
    \and
    Separation of content and style}

\end{abstract}

\section{Introduction}
\label{sec-introduction}

Recognizing text in the wild has attracted great interest in computer vision \citep{ye2015text,zhu2016scene,yang2017learning,shi2018aster,yang2019symmetry}. Recently, methods based on convolutional neural networks (CNNs) \citep{wang2012end,Jaderberg2015Deep,jaderberg2016reading} have significantly improved the accuracy of scene text recognition. Recurrent neural networks (RNNs) \citep{he2016reading,shi2016robust,shi2017end} and attention mechanism \citep{lee2016recursive,cheng2017focusing,cheng2017arbitrarily,yang2017learning} are also beneficial for recognition.

\begin{figure}[t]
\centering
\includegraphics[width=0.48\textwidth]{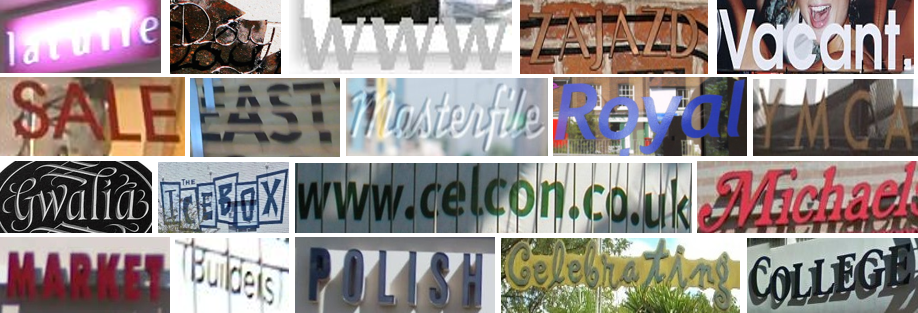}
\caption{Examples of scene text with complex backgrounds, making recognition very challenging.}
\label{fig:1-complex-background}
\end{figure}

Nevertheless, recognizing text in natural images is still challenging and largely remains unsolved \citep{shi2018aster}. As shown in Figure \ref{fig:1-complex-background}, text is found in various scenes, exhibiting complex backgrounds. The complex backgrounds cause difficulties for recognition. For instance, the complicated images often lead to attention drift \citep{cheng2017focusing} for attention networks. Thus, if the complex background style is normalized to a clean one, the recognition difficulty will significantly decreases.

\begin{figure}[t]
\centering
\includegraphics[width=.48\textwidth]{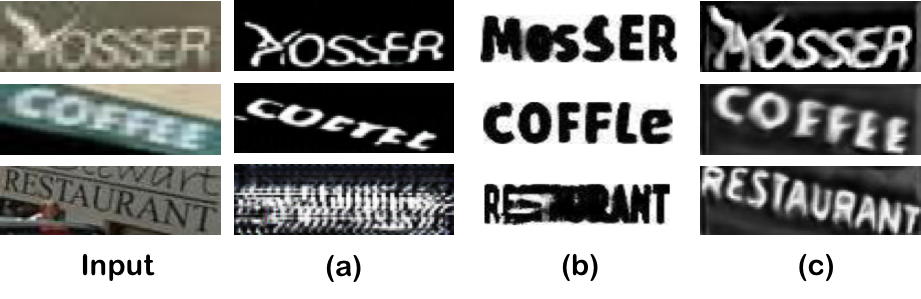}
\caption{Text content extraction of (a) CycleGAN, (b) Liu et al.\  \citep{liu2018synthetically} and (c) our method. Our method uses character-level adversarial training and thus better preserves the strokes of every character and removes complex backgrounds.}
\label{fig:cycleGAN}
\end{figure}

With the development of GANs \citep{Johnson2016hp,Cheng2019jc,Jing2019gf} in recent years, it is possible to migrate the scene background from a complex style to a clean style in scene text images. However, vanilla GANs are not sufficiently robust to generate sequence-like characters in natural images \citep{Fang2019vc}. As shown in Figure \ref{fig:cycleGAN} (a), directly applying the off-the-shelf CycleGAN fails to retain some strokes of the characters. In addition, as reported by Liu et al.\ \citep{liu2018synthetically}, applying a similar idea of image recovery to normalize the backgrounds for sequence-like objects fails to generate clean images. As illustrated in Figure \ref{fig:cycleGAN} (b), some characters on the generated images are corrupted, which leads to misclassification. One possible reason for this may be the discriminator is designed to focus on non-sequential object with a global coarse supervision \citep{zhang2019sequence}. Therefore, the generation of sequence-like characters requires more fine-grained supervision. 

One potential solution is to employ the pixel-wise supervision \citep{Isola2017kl}, which requires paired training samples aligning at pixel level. However, it is impossible to collect paired training samples in the wild. Furthermore, annotating scene text images with pixel-wise labels can be intractably expensive. To address the lack of paired data, it is possible to synthesize a large number of paired training samples, because synthetic data is cheaper to obtain. This may be why most state-of-the-art scene text recognition methods \citep{cheng2017arbitrarily,shi2018aster,cluo2019moran} only use synthetic samples \citep{jaderberg2014synthetic,gupta2016synthetic} for training, as tens of millions of training data are immediately available. However, experiments of Li et al.\  \citep{li2018show} suggest that there exists much room for improvement in synthesis engines. Typically a recognizer trained using real data significantly outperforms the ones trained using synthetic data due to the domain gap between artificial and real data. Thus, to enable broader application, our goal here is to improve GANs to meet the requirement of text image generation and address the unpaired data issue.

We propose an adversarial learning framework with an interactive joint training scheme, which achieves success in separating text content from background noises  by using only source training images and the corresponding text labels. The framework consists of an attention-based recognizer and a generative adversarial architecture. We take advantage of the attention mechanism in the attention-based recognizer to extract the features of each  character for further adversarial training. In contrast to global coarse supervisions, character-level adversarial training provides guidance for the generator in a fine-grained manner, which is critical to the success of our approach.

\begin{figure}[t]
\centering
\includegraphics[width=.48\textwidth]{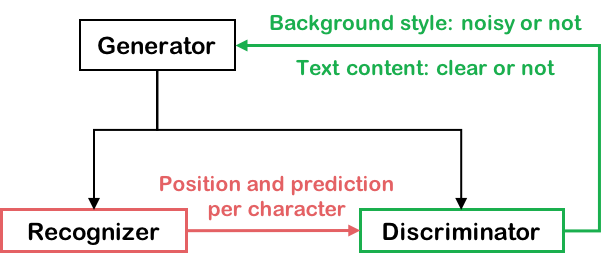}
\caption{Interactive joint training of our framework. The attention-based recognizer shares the position and prediction of every character with the discriminator, whereas the discriminator learns from the confusion of the recognizer, and guides the generator so that it can generate clear text content and clean background style to ease reading.}
\label{fig:overall-interactive}
\end{figure}

\begin{figure}[t]
\centering
\includegraphics[width=.48\textwidth]{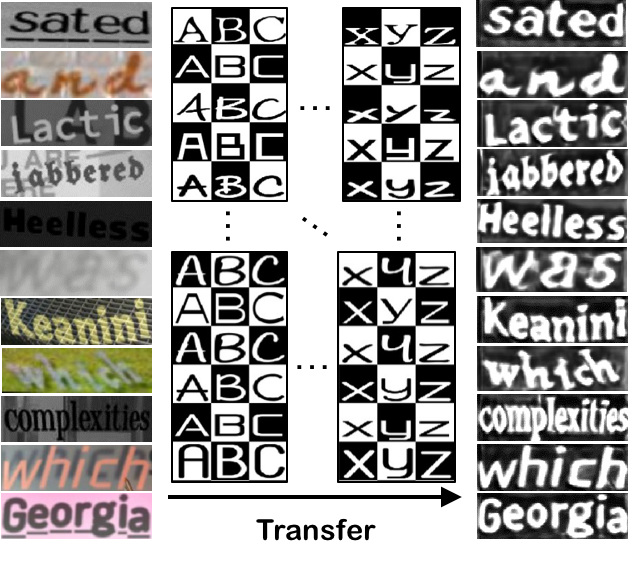}
\caption{Training samples and generations. Left: Widely used training datasets released by Jaderberg et al. \citep{jaderberg2014synthetic} and Gupta et al. \citep{gupta2016synthetic}. Middle: Unpaired target style samples, which are character-level and synthesized online. Right: Output of the generator.}
\label{fig:2-training-data}
\end{figure}

Our proposed framework is a meta framework. Thus, recent mainstream recognizers \citep{cheng2017arbitrarily,shi2018aster,cluo2019moran,li2018show} equipped with attention-based decoders \citep{bahdanau2014neural} can be integrated into our framework. As illustrated in Figure~\ref{fig:overall-interactive}, the attention-based recognizer predicts a mask for each character, which is shared with the discriminator. Thus, the discriminator is able to focus on every character and guide the generator to filter out various background styles while retaining  the character content. Benefiting from the advantage of the attention mechanism, the interactive joint training scheme requires only the images and corresponding text labels, without requirement of character bounding box annotation. Simultaneously, the target style training samples can be simply synthesized online during the training. As shown in Figure~\ref{fig:2-training-data}, for each target style sample, we randomly choose one character and simply render the character onto a clean background. Each sample contains a black character on a white background or a white character on a black background. The target style samples are character-level, whereas the input style samples are word-level. The unpaired training samples suggest our training process is flexible.

Moreover, we take a further step of the interactive joint training scheme. In addition to the sharing of attention masks, we proposed a feedback mechanism, which bridge the gap between the recognizer and the discriminator. The discriminator guides the generator according to the confusion of the recognizer. Thus, the erroneous character patterns on the generated images are corrected. For instance, the patterns of the characters ``C'' and ``G'' are similar, which can easily cause failed prediction of the recognizer. After the training using our feedback mechanism, the generated patterns are more discriminative, and incorrect predictions on ambiguous characters can be largely avoided.

To summarize, our main contributions are as follows.

\begin{itemize}

\item[$ 1) $] We propose a framework that separates text content from complex background styles to reduce recognition difficulty. The framework consists of an attention-based recognizer and a generative adversarial architecture. We devise an interactive joint training of them, which is critical to the success of our approach. 

\item[$ 2) $] The shared attention mask enables character-level adversarial training. Thus, the unpaired target style samples can be simply synthesized online. The training of our framework requires only the images and corresponding text labels. Additional annotations such as bounding boxes or pixel-wise labels are unnecessary. 

\item[$ 3) $] We further propose a feedback mechanism to improve the robustness of the generator. The discriminator learns from the confusion of the recognizer and guides the generator so that it can generate clear character patterns that facilitate reading.

\item[$ 4) $] Our experiments demonstrate that mainstream recognizers can benefit from our method and achieve new state-of-the-art performance by extracting text content from complex background styles. This suggests that our framework is a meta-framework, which is flexible for integration with recognizers.

\end{itemize}

\section{Related Work}
\label{section:Related work}

In this section, we review the previous methods that are most relevant to ours with respect to two categories: scene text recognition and generative adversarial networks.
 
\textbf{Scene text recognition.} Overviews of the notable work in the field of scene text detection and recognition have been provided by Ye et al.\  \citep{ye2015text} and Zhu et al.\  \citep{zhu2016scene}. The methods based on neural networks outperform the methods with hand crafted features, such as HOG descriptors \citep{dalal2005histograms}, connected components \citep{neumann2012real}, strokelet generation \citep{yao2014strokelets}, and label embedding \citep{rodriguez2015label}, because the trainable neural network is able to adapt to various scene styles. For instance, Bissacco et al. \citep{bissacco2013photoocr} applied a network with five hidden layers for character classification, and Jaderberg et al. \citep{Jaderberg2015Deep} proposed a CNN for unconstrained recognition. The CNN-based methods significantly improve the performance of recognition.

Moreover, the recognition models yield better robustness when they are integrated with RNNs \citep{he2016reading,shi2016robust,shi2017end} and attention mechanisms \citep{lee2016recursive,cheng2017focusing,cheng2017arbitrarily,yang2017learning}. For example, Shi et al.\  \citep{shi2017end} proposed an end-to-end trainable network using both CNNs and RNNs, namely CRNN. Lee et al.\  \citep{lee2016recursive} proposed a recursive recurrent network using attention modeling for scene text recognition. Cheng et al.\ \citep{cheng2017focusing} used a focusing attention network to correct attention alignment shifts caused by the complexity or low-quality of images. These methods have made great progress in regular scene text recognition.

With respect to irregular text, the irregular shapes introduce more background noise into the images, which increases recognition difficulty. To tackle this problem, Yang et al.\  \citep{yang2017learning} and Li et al.\  \citep{li2018show} used the two-dimensional (2D) attention mechanism for irregular text recognition. Liao et al.\  \citep{liao2019scene} recognized irregular scene text from a 2D perspective with a semantic segmentation network. Additionally, Liu et al.\  \citep{liu2016star}, Shi et al.\  \citep{shi2016robust,shi2018aster}, and Luo et al.\  \citep{cluo2019moran} proposed rectification networks to transform irregular text images into regular ones, which alleviates the interference of the background noise, and the rectified images become readable by a one-dimensional (1D) recognition network. Yang et al.\  \citep{yang2019symmetry} used character-level annotations for supervision for a more accurate description for rectification. Despite the many praiseworthy efforts that have been made, irregular scene text on complex backgrounds is still difficult to recognize in many cases.

\textbf{Generative adversarial networks.} With the widespread application of GANs \citep{goodfellow2014generative,mao2017least,odena2017conditional,Zhu2017hr}, font generation methods \citep{azadi2018multi,Yang2019Controllable} using adversarial learning have been successful on document images. These methods focus on the style of a single character and achieve incredible visual effects. 

However, our goal is to perform style normalization on noisy background, rather than the font, size or layout. A further challenge is to keep multiple characters for recognition. That means style normalization of the complex backgrounds of scene text images requires accurate separation between the text content and background noise. Traditional binarization/segmentation methods \citep{casey1996survey} typically work well on document images, but fail to handle the substantial variation in text appearance and the noise in natural images \citep{shi2018aster}. Style normalization of background in scene text images remains an open problem.

Recently, several attempts on  scene text generation have taken a crucial step forward. Liu et al.\  \citep{liu2018synthetically} guided the feature maps of an original image towards those of a clean image. The feature-level guidance reduces the recognition difficulty, whereas the image-level guidance does not result in a significant improvement in text recognition performance. Fang et al.\  \citep{Fang2019vc} designed a two-stage architecture to generate repeated characters in images. An additional 10k synthetic images boost the performance, but more synthetic images do not improve accuracy linearly. Wu et al.\ \citep{Wu2019bt} edited text in natural images using a set of corresponding synthetic training samples to preserve the style of both background and text. These methods provided sufficient visualized examples. However, the poor recognition performance on complex scene text remains a challenging problem.

We are interested in taking a further step to enable recognition performance to benefit from generation. Our method integrates the advantages of the attention mechanism and the GAN, and jointly optimizes them to achieve better performance. The text content is separated from various background styles, which are normalized for easier reading.

\section{Methodology}
\label{section:Methodology}
We design a framework to separate text content from noisy background styles, through an interactive joint training of an attention-based recognizer and a generative adversarial architecture. The shared attention masks from the attention-based recognizer enable character-level adversarial training. Then, the discriminator guides the generator to achieve background style normalization. In addition, a feedback mechanism bridges the gap between the discriminator and recognizer. The discriminator guides the generator according to the confusion of the recognizer. Thus, the generator can generate clear character patterns that facilitate reading.

In this section, we first introduce the attention decoder in mainstream recognizers. Then, we present a detailed description of the interactive joint training scheme. 

\subsection{Attention Decoder}

\begin{figure}[t]
\centering
\includegraphics[width=8cm,height=3.5cm]{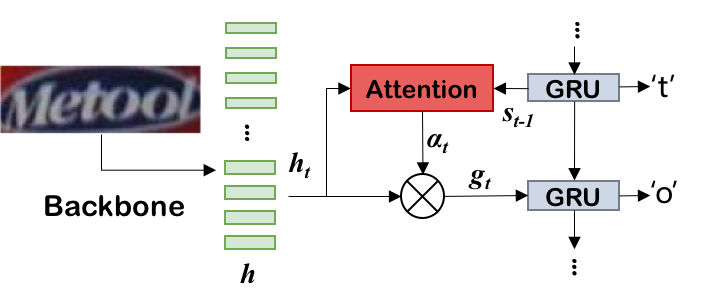}
\caption{Attention decoder, which recurrently attends to informative regions and outputs predictions.}
\label{fig:attention}
\end{figure}

\begin{figure*}[t]
\centering
\includegraphics[width=0.96\textwidth]{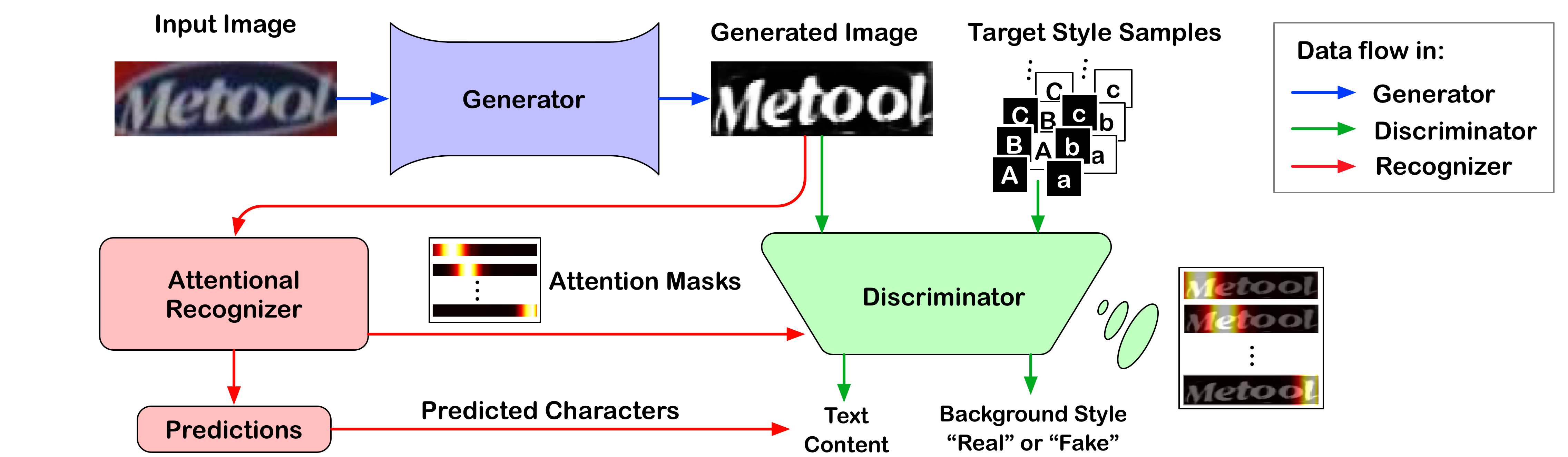}
\caption{Interactive joint training. The recognizer shares attention masks with the discriminator, whereas the discriminator learns from the predictions of the recognizer and updates the generator using ground truth. The shared attention masks work on feature maps, which we present in the generated images for better visualization.}
\label{fig:3-overview}
\end{figure*}

To date the attention decoder \citep{bahdanau2014neural} has become widely used in recent recognizers \citep{shi2018aster,cluo2019moran,li2018show,yang2019symmetry}. As shown in Figure \ref{fig:attention}, the decoder sequentially outputs predictions $(y_{1},y_{2} ...,y_{N})$ and stops processing when it predicts an end-of-sequence token $``EOS"$ \citep{sutskever2014sequence}. At time step $t$, output $y_t$ is given by
\begin{equation}
y_{t} = softmax(\bm{W}_{out}\bm{s}_{t}+b_{out}),
\end{equation}
where $\bm{s}_t$ is the hidden state at the $t$-th step. Then, we update $\bm{s}_t$ by
\begin{equation}
\bm{s}_t = GRU(\bm{s}_{t-1}, (y_{t-1}, \bm{g}_{t})),
\end{equation}
where $\bm{g}_{t}$ represents the glimpse vectors
\begin{equation}
\bm{g}_{t} = \sum_{i=1}^n(\alpha_{t,i} \bm{h}_{i}), \bm{\alpha}_{t} \in \mathbb{R}^n,
\label{equ-share}
\end{equation}
where $\bm{h}_{i}$ denotes the sequential feature vectors. Vector $\bm{\alpha}_{t}$ is the vector of attention mask, expressed as follows:
\begin{equation}
\alpha_{t,i} = \frac{\exp(e_{t,i})}{\sum_{j=1}^n(\exp(e_{t,j}))},
\end{equation}
\begin{equation}
e_{t,i} = \bm{w}^\mathrm{T}Tanh(\bm{W}_{s}\bm{s}_{t-1}+\bm{W}_{h}\bm{h}_{i}+b).
\end{equation}
Here, $\bm{W}_{out}$, $b_{out}$, $\bm{w}^\mathrm{T}$, $\bm{W}_{s}$, $\bm{W}_{h}$ and $b$ are trainable parameters. Note that $y_{t-1}$ is the $(t-1)$-th character in the ground truth in the training phase, whereas it is the previously predicted output in the testing phase. The training set is denoted as $\bm{D} = \left \{I_{i}, Y_{i} \right \}, i=1...N $. The optimization is to minimize the negative log-likelihood of the conditional probability of $\bm{D}$ as follows:
\begin{equation}
\mathcal{L}_{reg} = -\sum_{i=1}^N{ \sum_{t=1}^{| Y_{i} |}{\log p(Y_{i,t} \left| \right. I_{i}; \theta)} },
\end{equation}
where $Y_{i,t}$ is the ground truth of the $t$-th character in $I_{i}$ and $\theta$ denotes the parameters of the recognizer.

\subsection{Interactive Joint Training for Separating Text from Backgrounds}

As vanilla discriminator is designed for non-sequential object with a global coarse supervision, directly employing a discriminator fails to provide effective guidance for the generator. In contrast to apply a global discriminator, we supervise the generator in a fine-grained manner, namely, character-level adversarial learning, by taking advantage of the attention mechanism. Training the framework at character level also reduces the complexity of the preparation of target style data. Every target style sample containing one character can be easily synthesized online. 

\textbf{Sharing of attention masks.} Given an image $I$ as input, the goal of our generator $G$ is to generate a clean image $I'$ without a complex background. The discriminator $D$ encodes the image $I'$ as
\begin{equation}
\bm{E} = Encode(I').
\end{equation}
With similar settings of the backbone in recognizer (e.g., kernel size, stride size and padding size in the convolutional and pooling layers), the encoder in the discriminator is designed to output an embedding vector $\bm{E}_i$ with the same size as that of $\bm{h}_i$ in Equation \eqref{equ-share}, which enables the recognizer to share attention mask $\bm{\alpha}_{t}$ with the discriminator. After that, character-level features of the generation are extracted by
\begin{equation}
\bm{F}_{gen, t} = \sum_{i=1}^n(\alpha_{t,i} \bm{E}_{i}), \bm{\alpha}_{t} \in \mathbb{R}^n.
\end{equation}
The extracted character features are used for further adversarial training.

\textbf{Unpaired target style samples.} Benefiting from our character-level adversarial learning, target style samples can be simply synthesized online. As illustrated in Figures \ref{fig:2-training-data} and \ref{fig:3-overview}, every target style sample contains only a black character on a white background or a white character on a black background. The characters are randomly chosen. Following the previous methods for data synthesis \citep{jaderberg2014synthetic,gupta2016synthetic}, we collect fonts\footnote{\url{https://fonts.google.com}} to synthesize the target style samples. The renderer is a simple and publicly available engine\footnote{\url{https://pillow.readthedocs.io/en/stable/reference/ImageDraw.html}} that can efficiently synthesize samples online. Owing to the diversity of the fonts, the font sensitivity of the discriminator is thereby decreased, which enables the discriminator to focus on the background styles.

Because there is only one character in a target style image, we apply global average-pooling to the embedding features for every target style sample $I_{t}$ as follows:
\begin{equation}
\bm{F}_{tgt} = averagePooling\big(Encode(I_t)\big).
\end{equation}
The features of the $t$-th character in the generated image $\bm{F}_{gen, t}$ and the target style sample $\bm{F}_{tgt}$ are prepared for the following adversarial training.

\textbf{Adversarial training on style.} We use a style classifier in the discriminator to classify the style of characters in the generated images as fake and the characters in the target style samples as real. We use the 0$-$1 binary coding \citep{mao2017least} for style adversarial training, which is formulated as 
\begin{equation}
\small
\begin{split}
& \min \limits_{D} \mathcal{L}_{s}
 = \mathbb{E}_{I_t} [\big(1-Style(\bm{F}_{tgt})\big)^2]
+ \mathbb{E}_{I', t} \big[Style(\bm{F}_{gen, t})^2\big], \\
& \min \limits_{G} \mathcal{L}_{s} = \mathbb{E}_{I', t} \big[\big(1-Style(\bm{F}_{gen, t})\big)^2\big],
\end{split}
\end{equation}
where $Style(\cdot)$ denotes the style classifier.

The advantages of character-level adversarial training are threefold: 
1) Because the background is complicated in a scene text image, the background noise varies substantially in different character regions. Considering the text string as a whole and supervising the training in a global manner may cause the failure of the generator, as discussed previously in Section \ref{sec-introduction} and Figure \ref{fig:cycleGAN}. Thus, we encourage the discriminator to inspect the generation in a more fine-grained manner, namely, character-level supervision, which contributes to the effective learning. 
2) Training at character level brings a benefit for the preparation of target style data. For the synthesis of a text string, it is necessary to consider the text shape, the space between neighboring characters and the rotation of every character \citep{jaderberg2014synthetic,gupta2016synthetic}. In contrast, we can simply synthesize only one character on a clean background for every target style sample. Therefore, our target style samples can be simply synthesized online during the training.
3) The training is free of the need for paired data. Because the attention mechanism decomposes a text string into several characters and benefits the further training, only input scene text images and corresponding text labels are required. Hence, our framework is potentially flexible enough to make full use of available data to gain robustness.

\textbf{Feedback mechanism.} As our goal is to improve recognition performance, we are not only interested in the styles of the backgrounds, but also the quality of the generated content. Therefore, we use a content classifier in the discriminator to supervise content generation.

In contrast to the previous work auxiliary classifier GAN \citep{odena2017conditional}, which used ground truth to supervise the content classifier, our content classifier learns from the predictions of the recognizer. This bridges the gap between the recognizer and discriminator. The discriminator thus can guide the generator according to the confusion of the recognizer. After the training with this feedback mechanism, the generated patterns are more discriminative, which facilitates recognition. The details of the feedback mechanism are present as follows.

The generator $G$ and discriminator $D$ are updated by alternately optimizing
\begin{equation}
\begin{split}
\min \limits_{D} & \mathcal{L}_{c, D}=
\mathbb{E}_{(I, P), (I_t, GT)} [-\log Content(GT | \bm{F}_{tgt} ) \\
& -\frac{1}{|P|} \sum_{t=1}^{|P|} \log Content(P_t | \bm{F}_{gen, t})],
\label{eq:d-content}
\end{split}
\end{equation}

\begin{equation}
\begin{split}
\min \limits_{G}  \mathcal{L}_{c, G}=
\mathbb{E}_{I, GT}  [-\frac{1}{|GT|}\sum_{t=1}^{|GT|}\log  Content(GT_t | \bm{F}_{gen, t} )],
\end{split}
\end{equation}
where $GT$ denotes the ground truth of the input image $I$ and target style sample $I_t$. In addition, $Content(\cdot)$ is the content classifier. Note that the discriminator learns from the predictions $P$ on $I$ of the recognizer, whereas it uses $GT$ of $I$ to update the generator. This is an adversarial process that is similar to that of GAN training \citep{goodfellow2014generative,mao2017least,odena2017conditional,Zhu2017hr}. They use different labels for the discriminator and generator, but backpropagate the gradient using the same parameters as those of the discriminator. Alternately optimizing the discriminator and generator achieves adversarial learning.

There are some substitution errors in the predictions $P$ that are different from the $GT$. Therefore, the second term of the right side in Equation \eqref{eq:d-content} can be formulated as content adversarial training as
\begin{equation}
\begin{split}
-\frac{1}{|P|} & \sum_{t=1}^{|P|}\log Content(P_t|\bm{F}_{gen, t}) = \\
-\frac{1}{|P|} [ & \sum_{i=1}^{|P_{real}|}\log Content(P_{real, i}|\bm{F}_{gen, i}) \\
+ & \sum_{j=1}^{|P_{fake}|}\log Content(P_{fake, j}|\bm{F}_{gen, j}) ],
\end{split}
\end{equation}
where $P_{real}$ and $P_{fake}$ present the correct and incorrect predictions of the recognizer, respectively. Note that $P_{real} \cup P_{fake} = P$.

Since the discriminator with the content classifier learns from the predictions of the recognizer, it guides the generator to correct erroneous character patterns in the generated images. For instance, similar patterns such as ``C'' and ``G'', or ``O'' and ``Q", may cause failed prediction of the recognizer. If a ``G" is transformed to look more like a ``C" and the recognizer predicts it to be a ``C", the discriminator will learn that the pattern is a ``C" and guide the generator to generate a clearer ``G". We show more examples and further discuss this issue in Section \ref{section:Experiments}.

%
\def\varnothing{\mbox{\boldmath$ \emptyset$}}

\begin{algorithm}[b]
\SetAlgoLined
Discriminator: $D$; Generator: $G$\;
Batch size: $B$\;
Balance factor: $\beta$ (initialized as 1.0)\;
\While{not at the end of training}{
Sample $B$ training images as $I$, $I' = G(I)$\;
Randomly synthesize $B$ target style samples as $I_t$\;
Obtain the predictions $P$ on $I'$\;
Use $I'$ and $GT$ to update the recognizer: $\min \mathcal{L}_{reg}$, and obtain attention masks for the $D$\;
$I_{chosen} = \varnothing$\;
$P_{chosen} = \varnothing$\;
$GT_{chosen} = \varnothing$\;
\For{i in $1:B$}{
\If{$length(P_i) = length(GT_i)$}
{
\If{edit distance of $(P_i, GT_i)\le 1$}{
$I_{chosen} \leftarrow I_{chosen} \cup \{I'_i\}$\;
$P_{chosen} \leftarrow P_{chosen} \cup \{P_i\}$\;
$GT_{chosen} \leftarrow GT_{chosen} \cup \{GT_i\}$\;
}
}
}
\If{$I_{chosen} \ne \varnothing$}{
Generate a random number $k \in [0, 1)$\;
\If{$k \le \beta$}{
Use $I_{chosen}, P_{chosen}$ \\
to update the $D$: $\max \limits_{D} \mathcal{L}_{s}$,
$\min \limits_{D} \mathcal{L}_{c, D}$\;
}
Use $I_{chosen}, GT_{chosen}$ \\
to update the $G$: $\min \limits_{G} \mathcal{L}_{s}$,
$\min \limits_{G} \mathcal{L}_{c, G}$\;
$\beta \leftarrow \frac{\mathcal{L}_{s}+\mathcal{L}_{c, D}}{\mathcal{L}_{s}+\mathcal{L}_{c, G}}$\;
}
}
\caption{Interactive joint training}
\label{alg:interactive-training}
\end{algorithm}

\textbf{Interactive joint training.} The pseudocode of the interactive joint training scheme is presented in Algorithm \ref{alg:interactive-training}. During the training of our framework, we found that the discriminator often learns faster than the generator. A similar problem has also been reported by others \citep{BerthelotSM17,heusel2017gans}. The Wasserstein GAN \citep{ArjovskyCB17} uses  more update steps for the generator than the discriminator. We simply adjust the number of steps according to a balance factor $\beta \in (0, 1)$. If the discriminator learns faster than the generator, then the value of $\beta$ decreases, potentially resulting in a pause during the update steps for the discriminator. In practice, this trick contributes to the training stability of the generator.

We first sample a set of input samples, and randomly synthesize unpaired samples of target style. Then, the recognizer makes predictions on the generated images and shares its attention masks with the discriminator. To avoid the effects of incorrect alignment between character features and labels \citep{bai2018edit}, we filter out some predictions using the metrics of edit distance and string length. The corresponding images are also filtered out. Only substitution errors exist in the remaining predictions. Finally, the discriminator and generator are alternately optimized to achieve adversarial learning.

\begin{figure}[t]
\centering
\subfigure[]{
\begin{minipage}[c]{0.25\textwidth}
\centering
  \includegraphics[width=3.5cm, height=7cm]{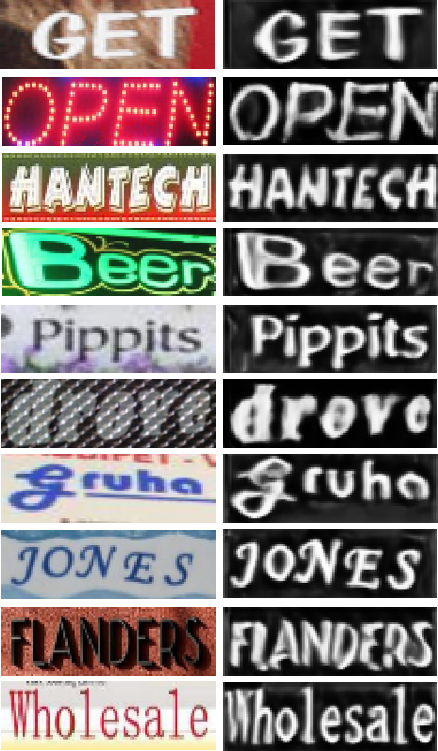}
\end{minipage}%
}%
\subfigure[]{
\begin{minipage}[c]{0.25\textwidth}
\centering
  \includegraphics[width=3.5cm, height=7cm]{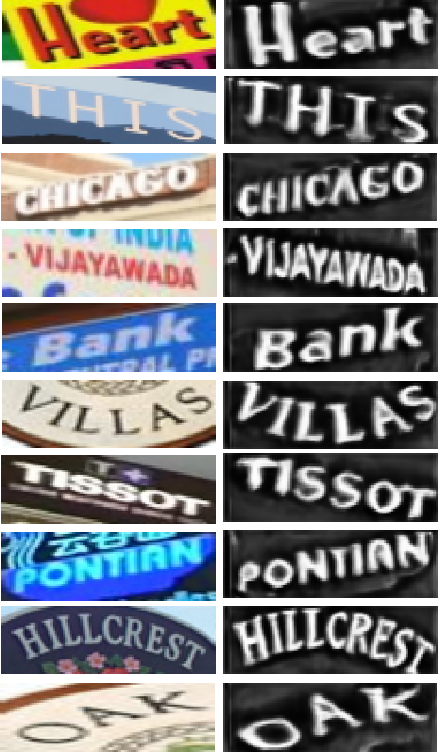}
\end{minipage}%
}%
\caption{Generated images for (a) regular and (b) irregular text.
Input images are on the left and the corresponding generated images are on the right. The text content is separated by the generator from the noisy background styles. In the generated images, the font style tends to be an average style.}
\label{fig:4-generated}
\end{figure}

After the adversarial training, the generator can separate text content from complex background styles. The generated patterns are clearer and easier to read. As illustrated in Figure \ref{fig:4-generated}, the generator works well on both regular text and slanted/curved text. Because the irregular shapes of the text introduce more surrounding background noise, the recognition difficulty can be significantly reduced by using our method.

\section{Experiments}
\label{section:Experiments}
In this section, we provide the training details and report the results of extensive experiments on various benchmarks, including both regular and irregular text datasets, demonstrating the effectiveness and generality of our method.

As paired text images in the wild are not available and there exists great diversity in the number of characters and image structure between the input images and our target style images, popular GAN metrics such as the inception score \citep{Salimans2016wg} and Fr\'echet inception distance \citep{heusel2017gans} cannot be directly applied in our evaluation. Instead, we use the word accuracy of recognition, which is a more straightforward metric, and is of interest for our target task, to measure the performance of all the methods. Recall that our goal here is to improve recognition accuracy.

\subsection{Datasets}

\textbf{SynthData}, which contains 6-million data released by Jaderberg et al. \citep{jaderberg2014synthetic} and 6-million data released by Gupta et al. \citep{gupta2016synthetic}, is a widely used training dataset. Following the most recent work for fair comparison, we select it as the training dataset. Only word-level labels are used, but other extra annotation is unnecessary in our framework. The model is trained using only synthetic text images, without any fine-tuning for each specific dataset.

IIIT5K-Words \citep{mishra2012scene} (\textbf{IIIT5K}) contains 3,000 cropped word images for testing. Every image has a 50-word lexicon and a 1,000-word lexicon. The lexicon consists of the ground truth and some randomly picked words.

Street View Text \citep{wang2011end} (\textbf{SVT}) was collected from the Google Street View, and consists of 647 word images. Each image is associated with a 50-word lexicon. Many images are severely corrupted by noise and blur or have very low resolutions. 

ICDAR 2003 \citep{lucas2003icdar} (\textbf{IC03}) contains 251 scene images that are labeled with text bounding boxes. For fair comparison, we discarded images that contain non-alphanumeric characters or those have fewer than three characters, following Wang et al. \citep{wang2011end}. The filtered dataset contains 867 cropped images. Lexicons comprise of a 50-word lexicon defined by Wang et al. \citep{wang2011end} and a “full lexicon”. The latter lexicon combines all lexicon words.

ICDAR 2013 \citep{karatzas2013icdar} (\textbf{IC13}) inherits most of its samples from IC03. It contains 1,015 cropped text images. No lexicon is associated with this dataset.

SVT-Perspective \citep{quy2013recognizing} (\textbf{SVT-P}) contains 645 cropped images for testing. Images were selected from side-view angle snapshots in Google Street View. Therefore, most images are perspective distorted. Each image is associated with a 50-word lexicon and a full lexicon.

CUTE80 \citep{risnumawan2014robust} (\textbf{CUTE}) contains 80 high-resolution images taken of natural scenes. It was specifically collected for evaluating the performance of curved text recognition. It contains 288 cropped natural images for testing. No lexicon is associated with this dataset.

ICDAR 2015 \citep{karatzas2015icdar} (\textbf{IC15}) contains 2077 images by cropping the words using the ground truth word bounding boxes. Cheng et al. \citep{cheng2017focusing} filtered out some extremely distorted images and used a small evaluation set (referred as \textbf{IC15-S}) containing only 1811 test images.

\begin{table*}[t]
\renewcommand\arraystretch{1.2}
\centering
\caption{Word accuracy on the testing datasets using different inputs. The recognizer is trained on the source and generated images, respectively.}
\label{table:ablation-study}
\setlength{\tabcolsep}{4mm}{
\begin{tabular}{c c c c c c c c }
\toprule
\multirow{2}{*}{Input Image} & \multicolumn{4}{c}{Regular Text} & \multicolumn{3}{c}{Irregular Text} \\
\cmidrule(lr){2-5} \cmidrule(lr){6-8}
& IIIT5K & SVT & IC03 & IC13 & SVT-P & CUTE & IC15 \\
\midrule
Source $I$ & 92.2 & 85.9 & 94.0 & 90.7 & 75.7 & 74.3 & 72.0 \\
Generation $I'$ & \textbf{92.5} & \textbf{86.6} & \textbf{95.0} & \textbf{91.4} & \textbf{79.2} & \textbf{80.9} & \textbf{73.0} \\
\bottomrule
\end{tabular}
}
\end{table*}

\subsection{Implementation Details}

As our proposed method is a meta-framework for recent attention-based recognition methods \citep{shi2018aster,cluo2019moran,li2018show,yang2019symmetry}, recent recognizers can be readily integrated with our framework. Thus the recognizer implementation follows their specific design. Here we present details of the discriminator, generator, and training.

\textbf{Generator. }The generator is a feature pyramid network (FPN)-like \citep{lin2017feature} architecture that consists of eight residual units. Each residual unit comprises a $1 \times 1$ convolution followed by two $3 \times 3$ convolutions. Feature maps are downsampled by $2 \times 2$ stride convolutions in the first three residual units. The numbers of output channels of the first four residual units are 64, 128, 256, and 256, respectively. The last four units are symmetrical with the first four, but we upsample the feature map by simple resizing. We apply element-wise addition to the output of the third and fifth units. At the top of the generator, there are two convolution layers that have 16 filters and one filter, respectively. 

\textbf{Discriminator. }The encoder in the discriminator consist of seven convolutional layers that have 16, 64, 128, 128, 192 and 256 filters. Their kernel sizes are all $3 \times 3$, except for the size of the last one, which is $2 \times 2$. The first, second, fourth and sixth convolutional layers are each followed by an average-pooling layer. Using settings similar to those of the backbone in the recognizer (e.g., kernel size, stride size and padding size in the convolutional and pooling layers), the output size of the encoder can be controlled to meet the requirements of the attention mask sharing of the recognizer. Both the style and content classifiers in the discriminator are one-layer fully connected networks.

\textbf{Training. }We use Adam \citep{kingma2015adam} to optimize the GAN. The learning rate is set to 0.002. It is decreased by a factor of 0.1 at epochs 2 and 4. In the interactive joint training, we utilize the attention mechanism in the recognizer. Therefore, an optimized attention decoder is necessary to enable the interaction. To accelerate the training process, we pre-trained the recognizer for three epochs.

\textbf{Implementation. }We implement our method using the PyTorch framework \citep{pytorch}. The target style samples are resized to $32 \times 32$. Input images are resized to $64 \times 256$ for the generator and $32 \times 100$ for the recognizer. The outputs of the generator are also resized to $32 \times 100$. When the batch size is set to 64, the training speed is approximately 1.7 iterations/sec. Our method takes an average of 1.1 ms to generate an image using an NVIDIA GTX-1080Ti GPU.


\subsection{Ablation Study}

\textbf{Experiment setup.} To investigate the effectiveness of separating text content from noisy background styles, we conduct an ablation analysis by using a simple recognizer. The backbone of the recognizer is a 45-layer residual network \citep{he2016deep}, which is a popular architecture \citep{shi2018aster}. On the top of the backbone, there is an attention-based decoder. In the decoder, the number of GRU hidden units is $256$. The decoder outputs 37 classes, including 26 letters, 10 digits, and a symbol that represented $``\rm EoS"$. The training data is SynthData. We evaluate the recognizer on seven benchmarks, including regular and irregular text.

\textbf{Input of the recognizer.} 
We study the contribution of our method by replacing the generated image with the corresponding input image. The results are listed in Table~\ref{table:ablation-study}. The recognizer trained using SynthData serves as a baseline.
Compared to the baseline, the clean images generated by our method boost recognition performance. We observe that the improvement is more substantial on irregular text. One notable improvement is an accuracy increase of 6.6\% on CUTE. One possible reason for this is that the irregular text shapes introduce more background noise than the regular ones. Because our method removes the surrounding noise and extracts the text content for recognition, the recognizer can thus focus on characters and avoid noisy interference. 

With respect to regular text, the baseline is much higher and there is less room for improvement, but our method also shows advantages in recognition performance. The gain of performance on several kinds of scene text, including low quality images in SVT and real scene images in IC03/IC13, suggests the generality our method. To summarize, the generated clean images of our proposed method greatly decrease recognition difficulty.

\begin{figure}[t]
\centering
\includegraphics[width=7cm,height=1cm]{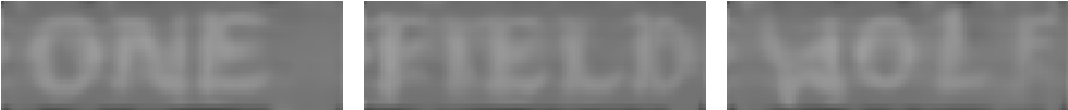}
\caption{Visualization of background normalization weakly supervised by content label.}
\label{fig:weak}
\end{figure}

\textbf{Style supervision.} We study the necessity of style supervision by disabling the style classifier in the discriminator. Without style adversarial training, the background style normalization is only weakly supervised by the content label. As shown in the Figure \ref{fig:weak}, the generated images suffer from severe image degradation, which leads to poor robustness of the recognizer. The quantitative recognition results of not using/using the style supervision are presented in the second and third row of Table \ref{table:ablation-study2}. The significant gaps indicate that without the style supervision, the quality of the generated images is insufficient for recognition training. Thus, the style adversarial training is necessary and is used in the basic design of our method.

\begin{table*}[t]
\renewcommand\arraystretch{1.2}
\centering
\caption{Word accuracy on generated images using variants of content supervision for the discriminator. Losses $\mathcal{L}_{s}$ and $\mathcal{L}_{c}$ denote style loss and content loss, respectively.}
\label{table:ablation-study2}
\setlength{\tabcolsep}{3.5mm}{
\begin{tabular}{ c c c c c c c }
\toprule
\multirow{2}{*}{Content Supervision} & \multicolumn{2}{c}{Adversarial Loss} &  \multirow{2}{*}{Feedback Mechanism} & \multicolumn{3}{c}{Testing Set} \\
\cmidrule(lr){2-3}\cmidrule(lr){5-7}
& $\mathcal{L}_{s}$ & $\mathcal{L}_{c}$ &  & SVT-P & CUTE & IC15 \\
\midrule
None & \checkmark & $\times$ & $\times$ & Failed & Failed & Failed \\
Ground Truth & $\times$ & \checkmark & $\times$ & 71.9 & 74.7 & 65.5 \\
Ground Truth & \checkmark & \checkmark & $\times$ & 75.0 & 78.4 & 72.5 \\
Prediction of Recognizer & \checkmark & \checkmark & \checkmark & \textbf{79.2} & \textbf{80.9} & \textbf{73.0} \\
\bottomrule
\end{tabular}
}
\end{table*}

\textbf{Feedback mechanism.} We also study the effectiveness of the content classifier in the discriminator and the proposed feedback mechanism. In this experiment, we first disable the content classifier. Therefore, there is no content supervision. Only a style adversarial loss supervises the generator. The result is shown in the first row in Table \ref{table:ablation-study2}. The accuracy on the generated images decreases to nearly zero. We observe that the generator fails to retain the character patterns for recognition. As the content classifier is designed  for assessing the discriminability and diversity of samples \citep{odena2017conditional}, it is important to guide the generator so that it can determine informative character patterns and retain them for recognition. When the content supervision is not available, the generator is easily trapped into failure modes, namely mode collapse \citep{Salimans2016wg}. Therefore, the content supervision in the discriminator is necessary.

Then we enable the content classifier and replace the supervision in $\mathcal{L}_{c}$ with the ground truth. This setting is similar to that of  the auxiliary classifier GANs \citep{odena2017conditional}, which use content supervision for discriminability and diversity in the style adversarial training. After this process, the generated text images contain text content for recognition. 

Finally, we replace the content supervision with the predictions of the recognizer. The discriminator thus learns from the confusion of the recognizer, and guides the generator so that it can refine the character patterns to be easier to read. Therefore, the adversarial training is more relevant to the recognition performance. As shown in Table \ref{table:ablation-study2}, the feedback mechanism further improves the robustness of the generator and benefits the recognition performance. 

\begin{table*}[t]
\renewcommand\arraystretch{1.2}
\centering
\caption{Word accuracy on testing datasets using different transformation methods.}
\label{table:style-transfer-study}
\setlength{\tabcolsep}{3.3mm}{
\begin{tabular}{c c c c c c c c c}
\toprule
\multirow{2}{*}{Transformation} & \multirow{2}{*}{Method} & \multicolumn{4}{c}{Regular Text} & \multicolumn{3}{c}{Irregular Text} \\
\cmidrule(lr){3-6} \cmidrule(lr){7-9}
& & IIIT5K & SVT & IC03 & IC13 & SVT-P & CUTE & IC15 \\
\midrule
\multirow{3}{*}{Style Normalization} & OTSU \citep{otsu1979threshold} & 70.3 & 65.4 & 76.0 & 76.0 & 46.5 & 50.3 & 49.3 \\
& CycleGAN \citep{Zhu2017hr} & 43.6 & 21.3 & 37.0 & 35.9 & 14.6 & 18.8 & 18.0 \\
& Ours & {92.5} & 86.6 & 95.0 & 91.4 & 79.2 & 80.9 & 73.0 \\
\midrule
\multirow{2}{*}{+ Rectification} & Ours + ASTER \citep{shi2018aster} & 94.0 & 90.0 & 95.6 & 93.3 & 81.6 & 85.1 & 78.1 \\
& Ours + ESIR \citep{zhan2019esir} & \textbf{94.1} & \textbf{90.6} & \textbf{96.0} & \textbf{94.2} & \textbf{82.2} & \textbf{87.8} & \textbf{78.5} \\
\bottomrule
\end{tabular}
}
\end{table*}

\begin{figure}[t]
\centering
\includegraphics[width=8.5cm,height=4.5cm]{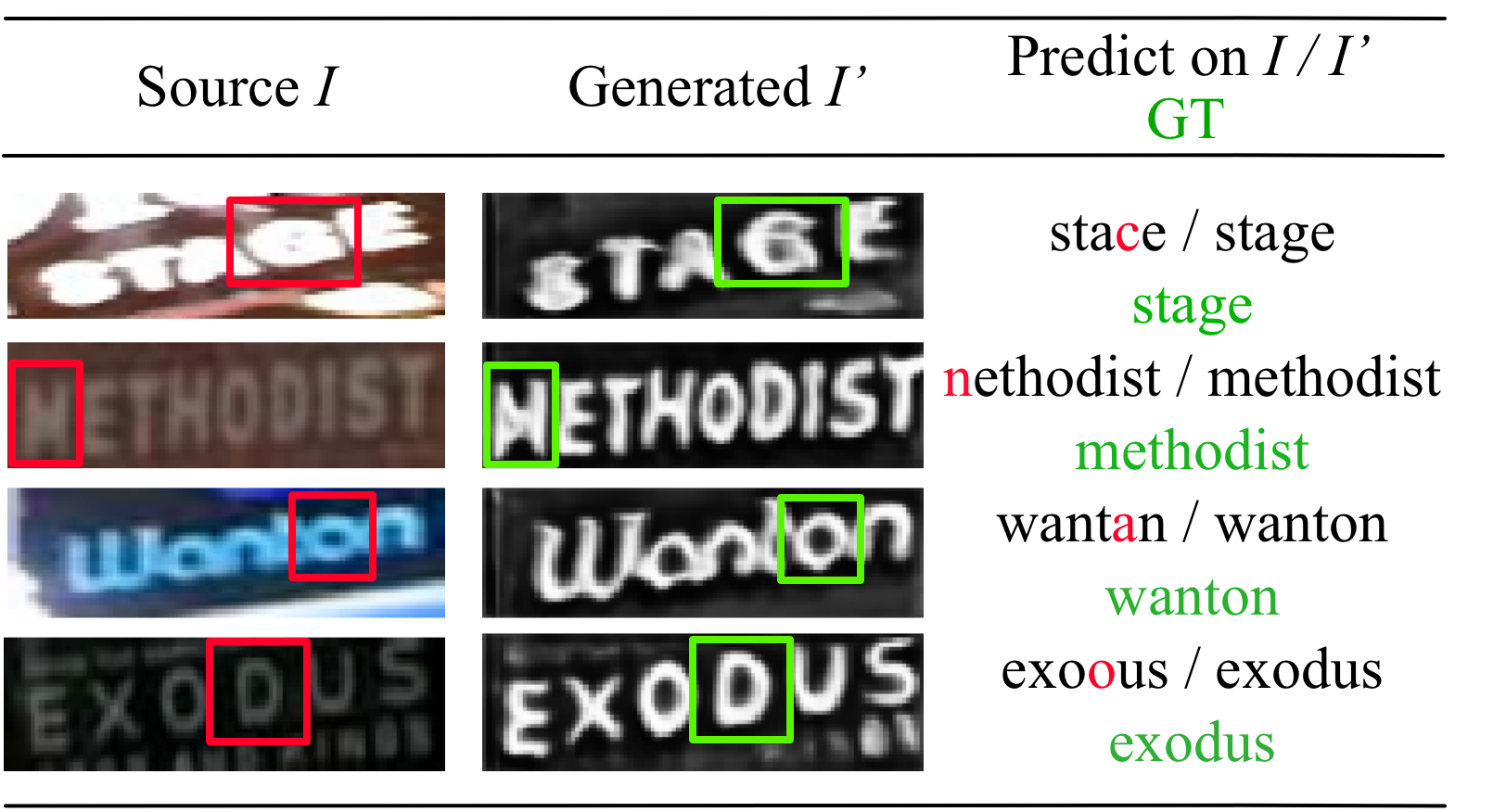}
\caption{Predictions of challenging samples in the SVT-P testing set. Recognition errors are marked as red characters. Confusing and distinct patterns are marked by red and green bounding boxes, respectively.}
\label{fig:ablation-study}
\end{figure}

One interesting observation is that on the SVT-P testing set, the accuracy on the source image (75.7\% in Table \ref{table:ablation-study}) is higher than that on the generated image with content supervision of the ground truth (75.0\% in Table \ref{table:ablation-study2}). We observe the source samples and find that most images are severely corrupted by noise and blur. Some of them have low resolutions. The characters in the generated samples are also difficult to distinguish. After training with the feedback mechanism, the generator is able to generate clear patterns that facilitate reading, which boosts the recognition accuracy from 75.0\% to 79.2\%. As illustrated in Figure \ref{fig:ablation-study}, the predictions of ``C" and ``N" are corrected to ``G" and ``M", respectively. The clear characters in the generated images are easier to read.

\subsection{Comparisons with Generation Methods}

Recently, a large body of literature \citep{shi2018aster,cluo2019moran,li2018show,yang2019symmetry} has explored the use of stronger recognizers to tackle the complications in scene text recognition. However, there is little consideration of the quality of the source images. The  background noise in the source image has not been addressed intensively before. To the best of our knowledge, our method may be the first image generation network that removes background noise and retains text content to benefit recognition performance. Although few literature proposed to address this issue stated above, we find several popular generation methods and perform comparisons under fair experimental conditions. A pre-trained recognizer used in the ablation study is adopted in the comparisons. The recognizer is then be fine-tuned on different kinds of generations. 

\begin{figure}[t]
\centering
\includegraphics[width=7cm,height=5.5cm]{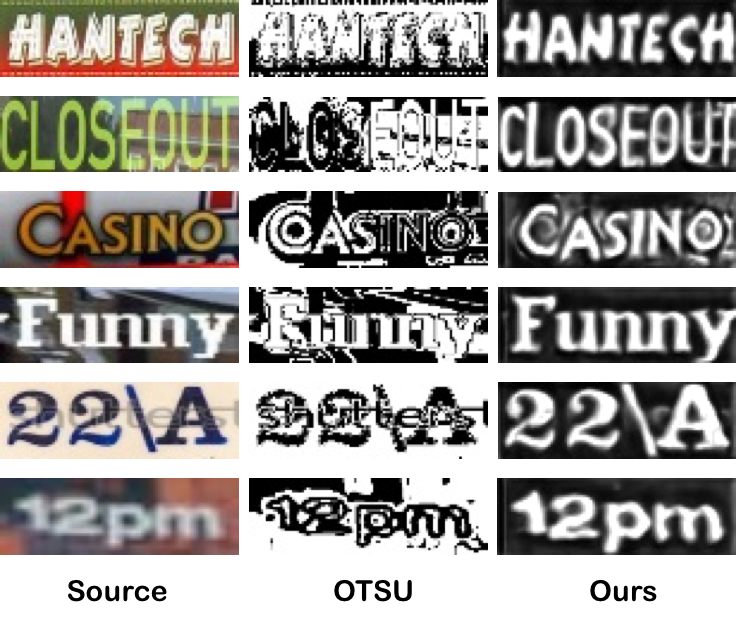}
\caption{Comparison between the OTSU \citep{otsu1979threshold} and our method.}
\label{fig:binary}
\end{figure} 

\begin{table*}[t]
\renewcommand\arraystretch{1.12}
\centering
\caption{Word accuracy on regular benchmarks. ``50", ``1k" and ``0" are lexicon sizes. ``Full" indicates the combined lexicon of all images in the benchmarks. ``Add." means the method uses extra annotations, such as character-level bounding boxes and pixel-level annotations. ``Com." is the proposed ensemble method that outputs prediction of either source or generated image with higher confidence score.}
\label{table:Results on general benchmarks}
\setlength{\tabcolsep}{3.5mm}{
\begin{tabular}{r  c c c c c c c c c c}
\toprule
\multirow{2}{*}{Method} & \multirow{2}{*}{Add.} & \multicolumn{3}{c}{IIIT5K} & \multicolumn{2}{c}{SVT} &\multicolumn{3}{c}{IC03} & IC13 \\
\cmidrule(lr){3-5}\cmidrule(lr){6-7}\cmidrule(lr){8-10}\cmidrule(lr){11-11} & & 50 & 1k & 0 & 50 & 0 & 50 & Full & 0 & 0\\
\midrule
\citet{yao2014strokelets} & & 80.2 & 69.3 & - & 75.9 & - & 88.5 & 80.3 & - & - \\
\citet{jaderberg2014deep} & & - & - & - & 86.1 & - & 96.2 & 91.5 & - & -\\
\citet{su2014accurate} & & - & - & - & 83.0 & - & 92.0 & 82.0 & - & - \\
\citet{rodriguez2015label} & & 76.1 & 57.4 & - & 70.0 & - & - & - & - & - \\
\citet{gordo2015supervised} & & 93.3 & 86.6 & - & 91.8 & - & - & - & - & - \\
\citet{Jaderberg2015Deep} & & 95.5 & 89.6 & - & 93.2 & 71.7 & 97.8 & 97.0 & 89.6 & 81.8 \\
\citet{jaderberg2016reading} & & 97.1 & 92.7 & - & 95.4 & 80.7 & 98.7 & 98.6 & 93.1 & 90.8 \\
\citet{shi2016robust} & & 96.2 & 93.8 & 81.9 & 95.5 & 81.9 & 98.3 & 96.2 & 90.1 & 88.6 \\
\citet{lee2016recursive} & & 96.8 & 94.4 & 78.4 & 96.3 & 80.7 & 97.9 & 97.0 & 88.7 & 90.0\\
\citet{liu2016star} & & 97.7 & 94.5 & 83.3 & 95.5 & 83.6 & 96.9 & 95.3 & 89.9 & 89.1 \\
\citet{shi2017end} & & 97.8 & 95.0 & 81.2 & 97.5 & 82.7 & 98.7 & 98.0 & 91.9 & 89.6 \\
\citet{yang2017learning} & \checkmark & 97.8 & 96.1 & - & 95.2 & - & 97.7 & - & - & -\\
\citet{cheng2017focusing} & & 98.9 & 96.8 & 83.7 & 95.7 & 82.2 & 98.5 & 96.7 & 91.5 & 89.4 \\
\citet{liu2018char} & \checkmark & - & - & 83.6 & - & 84.4 & - & - & 91.5 & 90.8 \\
\citet{liu2018synthetically} & & 97.3 & 96.1 & 89.4 & 96.8 & 87.1 & 98.1 & 97.5 & 94.7 & 94.0 \\
\citet{liu2018squeezedtext} & \checkmark & 97.0 & 94.1 & 87.0 & 95.2 & - & 98.8 & 97.9 & 93.1 & 92.9 \\
\citet{cheng2017arbitrarily} & & 99.6 & 98.1 & 87.0 & 96.0 & 82.8 & 98.5 & 97.1 & 91.5 & - \\
\citet{bai2018edit} & & 99.5 & 97.9 & 88.3 & 96.6 & 87.5 & 98.7 & 97.9 & 94.6 & 94.4\\
\citet{shi2018aster} & & 99.6 & \textbf{98.8} & 93.4 & 97.4$^{\rm \color{red}{1}}$ & 89.5$^{\rm \color{red}{1}}$ & 98.8 & 98.0 & 94.5 & 91.8 \\
\citet{cluo2019moran} & & 97.9 & 96.2 & 91.2 & 96.6 & 88.3 & 98.7 & 97.8 & 95.0 & 92.4 \\
\citet{liao2019scene} & \checkmark  & \textbf{99.8} & \textbf{98.8} & 91.9 & 98.8 & 86.4 & - & - & - & 91.5 \\
\citet{li2018show} & & - & - & 91.5 & - & 84.5 & - & - & - & 91.0                      \\
\citet{zhan2019esir} & & 99.6 & \textbf{98.8} & 93.3 & 97.4 & 90.2 & - & - & - & 91.3  \\
\citet{yang2019symmetry} & \checkmark & 99.5 & \textbf{98.8} & 94.4 & 97.2 & 88.9 & 99.0 & 98.3 & 95.0 & 93.9 \\
\midrule
ASTER & & 99.1 & 97.9 & 93.5 & 98.0 & 88.6 & 98.8 & 98.0 & 94.7 & 92.0 \\
+ Ours & & 99.1 & 98.0 & 94.0 & 98.3 & 90.0 & 98.8 & 98.1 & 95.6 & 93.3 \\
+ Com. & & 99.6 & 98.7 & 95.4 & 98.9 & 92.7 & \textbf{99.1} & \textbf{98.8} & \textbf{96.3} & 94.8 \\
\midrule
ESIR & & 99.2 & 98.0 & 93.8 & 98.0 & 88.7 & 98.8 & 98.2 & 95.0 & 93.5 \\
+ Ours & & 99.5 & 98.6 & 94.1 & 98.0 & 90.6 & 98.8 & 98.5 & 96.0 & 94.2 \\
+ Com. & & 99.6 & \textbf{98.8} & \textbf{95.6} & \textbf{99.4} & \textbf{92.9} & \textbf{99.1} & \textbf{98.8} & 96.2 & \textbf{96.0} \\
\bottomrule
\end{tabular}
}
\begin{tablenotes}
\item \footnotesize $^{\rm \color{red}{1}}$The result was corrected by the authors on \url{https://github.com/bgshih/aster}.
\end{tablenotes}
\end{table*}

First we use a popular binarization method, namely OTSU method \citep{otsu1979threshold}, to separate the text content from the background noise by binarizing the source images. As shown in Figure \ref{fig:binary}, we visualize the binarized images and find that single threshold value is not sufficiently robust to separate the foreground and background in scene text images, because the background noise usually follows multimodal distribution. Therefore, the recognition accuracy on the generation of OTSU method falls behind ours in Table~\ref{table:style-transfer-study}.
\begin{figure}[t]
\centering
\includegraphics[width=0.425\textwidth]{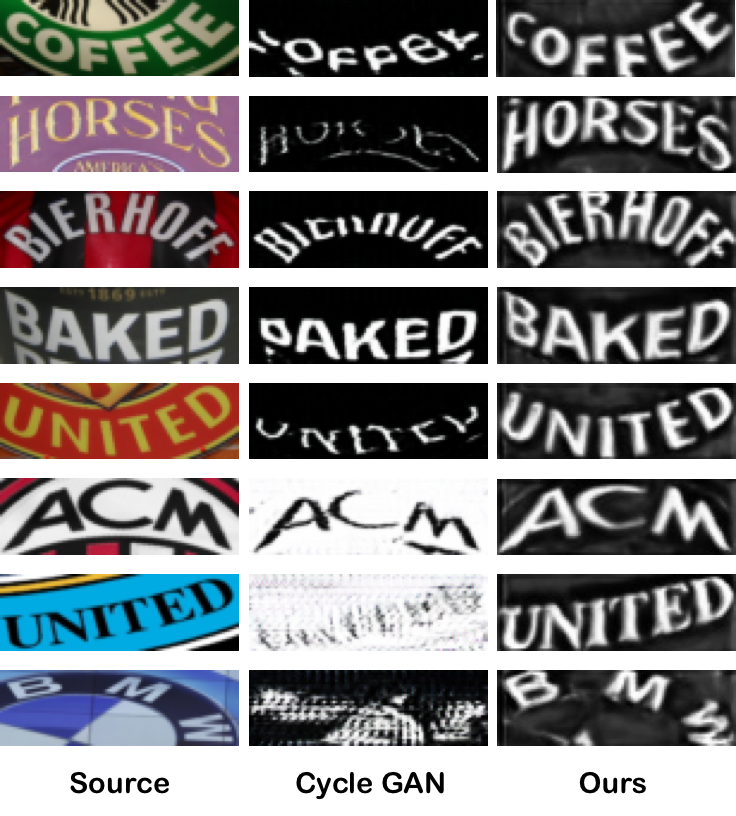}
\caption{Comparison between 
CycleGAN \citep{Zhu2017hr} and our method.}
\label{fig:cycle}
\end{figure}

\begin{table*}[t]
\renewcommand\arraystretch{1.15}
\centering
\caption{Word accuracy on irregular benchmarks. ``50" and ``0" are lexicon sizes. ``Full" indicates the combined lexicon of all images in the benchmarks. ``Add." means the method uses extra annotations, such as character-level bounding boxes and pixel-level annotations. ``Com." is the proposed ensemble method that outputs prediction of either source or generated image with higher confidence score.}
\label{table:Results on Irregular}
\setlength{\tabcolsep}{4.5mm}
{
\begin{tabular}{ r  c c c c c c c }
\toprule
\multirow{2}{*}{Method} & \multirow{2}{*}{Add.} & \multicolumn{3}{c}{SVT-P} & CUTE & IC15-S & IC15 \\
\cmidrule(lr){3-5}\cmidrule(lr){6-6}\cmidrule(lr){7-7}\cmidrule(lr){8-8}
& & 50 & Full & 0 & 0 & 0 & 0\\
\midrule
\citet{shi2016robust} & & 91.2 & 77.4 & 71.8 & 59.2 & - & - \\
\citet{liu2016star} & & 94.3 & 83.6 & 73.5 & - & - & - \\
\citet{shi2017end} & & 92.6 & 72.6 & 66.8 & 54.9 & - & - \\
\citet{yang2017learning} & \checkmark & 93.0 & 80.2 & 75.8 & 69.3 & -& - \\
\citet{cheng2017focusing} & & 92.6 & 81.6 & 71.5 & 63.9 & 70.6 & - \\
\citet{liu2018char} & \checkmark & - & - & - & - & - & 60.0 \\
\citet{liu2018synthetically} & & - & - & 73.9 & 62.5 & - & - \\
\citet{cheng2017arbitrarily} & & 94.0 & 83.7 & 73.0 & 76.8 & - & 68.2 \\
\citet{bai2018edit} & & - & - & - & - & 73.9 & - \\
\citet{shi2018aster} & & - & - & 78.5 & 79.5 & 76.1 & - \\
\citet{cluo2019moran} & & 94.3 & 86.7 & 76.1 & 77.4 & - & 68.8 \\
\citet{liao2019scene} & \checkmark & - & - & - & 79.9 & - & - \\
\citet{li2018show} & & - & - & 76.4 & 83.3 & - & 69.2 \\
\citet{zhan2019esir} & & - & - & 79.6 & 83.3 & - & 76.9 \\
\citet{yang2019symmetry} & \checkmark & - & - & 80.8 & 87.5 & - & 78.7 \\
\midrule
ASTER & & 94.3 & 87.3 & 77.7 & 79.9 & 75.8 & 74.0\\
+ Ours & & 95.0$^{\color{red}{\uparrow0.7}}$ & 90.1$^{\color{red}{\uparrow2.8}}$ & 81.6$^{\color{red}{\uparrow3.9}}$ & 85.1$^{\color{red}{\uparrow5.2}}$ & 80.1$^{\color{red}{\uparrow4.3}}$ & 78.1$^{\color{red}{\uparrow4.1}}$ \\
+ Com. & & 95.5 & \textbf{92.2} & \textbf{85.4} & 89.6 & 83.7 & 81.1 \\
\midrule
ESIR & & 94.3 & 87.3 & 79.8 & 83.7 & 79.3 & 77.1\\
+ Ours & & 95.0$^{\color{red}{\uparrow0.7}}$ & 89.3$^{\color{red}{\uparrow2.0}}$ & 82.2$^{\color{red}{\uparrow2.4}}$ & 87.8$^{\color{red}{\uparrow4.1}}$ & 81.1$^{\color{red}{\uparrow1.8}}$ & 78.5$^{\color{red}{\uparrow1.4}}$ \\
+ Com. & & \textbf{95.8} & 91.5 & 85.1 & \textbf{91.3} & \textbf{83.9} & \textbf{81.4} \\
\bottomrule
\end{tabular}
}
\end{table*}

Then, we compare our method with generation methods. Considering the high demand for data (pixel-level paired samples) of pixel-to-pixel GANs \citep{Isola2017kl}, we treat this kind of method as a potential solution when there is no restriction of data. Here, we study the CycleGAN\footnote{The official implementation is available on \url{https://github.com/junyanz/pytorch-CycleGAN-and-pix2pix}}
\citep{Zhu2017hr}. Before the training, we synthesize word-level clean images as target style samples. The results shown in Table \ref{table:style-transfer-study} and Figure \ref{fig:cycle} suggest that modeling a text string with multiple characters as a whole leads to poor retention of character details. The last two rows in Figure \ref{fig:cycle} are failed generations, which indicate that the generator fails to model the relationships of the characters. In Table \ref{table:style-transfer-study}, the recognition accuracy on this kind of generation drops substantially.

Compared with previous methods, our method not only normalizes noisy backgrounds to a clean style, but also generates clear character patterns that tend to be an average style. The end-to-end training with the feedback mechanism benefits the recognition performance. 
We also show the effectiveness of the image rectification by integrating our method with advanced rectification modules \citep{shi2018aster,zhan2019esir}. It can be seen that image rectifiers are still significant for improving recognition performance. Thus, different from irregular text shape, noisy background style is another challenge.

\subsection{Integration with State-of-the-art Recognizers}

As our method is a meta-framework, it can be integrated with recent recognizers \citep{cheng2017arbitrarily,shi2018aster,cluo2019moran,li2018show,yang2019symmetry} equipped with attention-based decoders \citep{bahdanau2014neural}. We conduct experiments using representative methods, namely ASTER \citep{shi2018aster} and ESIR \citep{zhan2019esir}, to investigate the effectiveness of our framework. The reimplementation results are comparable with those in the paper. With respect to the dataset providing a lexicon, we choose the lexicon word under the metric of edit distance. The results of comparison with previous methods are shown in Tables \ref{table:Results on general benchmarks} and \ref{table:Results on Irregular}. All the results of the previous methods are collected from their original papers. If a method uses extra annotations, such as character-level bounding boxes and pixel-level annotations, we indicate this with ``Add.". For fair comparison, we perform a comparison with the method of Li et al. \citep{li2018show} by including the results of their model trained using synthetic data. 

\begin{figure}[b]
\centering
\includegraphics[width=6.5cm,height=5cm]{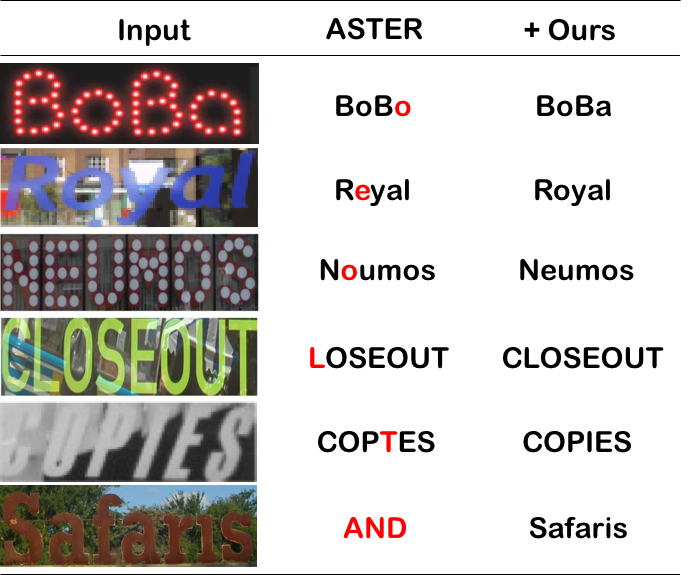}
\caption{Predictions corrected by our method.}
\label{fig:7-comparison}
\end{figure} 

Using the strong baseline of ASTER, we first evaluate the contribution of our method on regular text as shown in Table \ref{table:Results on general benchmarks}. Although the baseline accuracy on these benchmarks is high, thus no much room for improvement, our method still achieves a notable improvement in lexicon-free prediction. For instance, it leads to accuracy increases of 1.4\% on SVT and 1.3\% on IC13. Some predictions corrected using our generations are shown in Figure \ref{fig:7-comparison}. Then, we reveal the superiority of our method by applying it to irregular text recognition. As shown in Table \ref{table:Results on Irregular}, our method significantly boosts the performance of ASTER by generating clean images. The ASTER integrated with our approach outperforms the baseline by a wide margin on SVT (3.9\%), CUTE (5.2\%) and IC15 (4.3\%). This suggests that our generator removes the background noise introduced by irregular shapes and further reduces difficulty of rectification and recognition. It is noteworthy that the ASTER with our method outperforms ESIR \citep{zhan2019esir} that uses more rectification iterations (ASTER only rectifies the image once), which demonstrates the significant contribution of our method. The performance is even comparable with the state-of-the-art method \citep{yang2019symmetry}, which uses character-level geometric descriptors for supervision. Our method achieves a better trade-off between recognition performance and data requirement.

After that, our method is integrated with a different method ESIR to show the generalization. Based on the more advanced recognizer, our method can achieve further gains. For instance, the improvement is still notable on CUTE (4.1\%). As a result, the performance of the ESIR is also significantly boost by our method.

\begin{figure}[t]
\centering
\includegraphics[width=8cm,height=2.5cm]{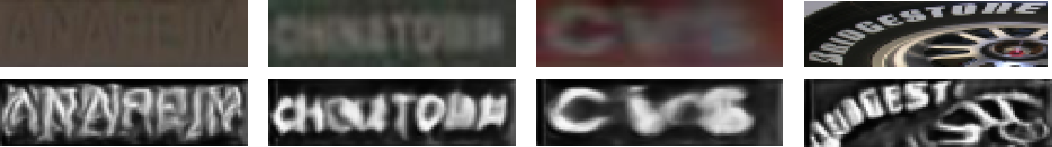}
\caption{Failure cases. Top: source images. Bottom: generated images.}
\label{fig:failure}
\end{figure} 

\begin{table*}[t]
\renewcommand\arraystretch{1.15}
\centering
\caption{Word accuracy on testing datasets when we use a little more real training data.}
\label{table:real-data}
\setlength{\tabcolsep}{3.5mm}{
\begin{tabular}{c c c c c c c c c c}
\toprule
\multirow{2}{*}{Method} & \multirow{2}{*}{Training data} & \multicolumn{4}{c}{Regular Text} & \multicolumn{4}{c}{Irregular Text} \\
\cmidrule(lr){3-6} \cmidrule(lr){7-10}
& & IIIT5K & SVT & IC03 & IC13 & SVT-P & CUTE & IC15-S & IC15 \\
\midrule
\multirow{2}{*}{ASTER} & Millions of Synthetic Data & 93.5 & 88.6 & 94.7 & 92.0 & 77.7 & 79.9 & 75.8 & 74.0 \\
& + 50k Real Data & 94.5 & 90.4 & 95.0 & 92.9 & 79.4 & 88.9 & 84.1 & 81.3 \\
\midrule
\multirow{2}{*}{+ Ours} & Millions of Synthetic Data & 95.4 & 92.7 & \textbf{96.3} & 94.8 & 85.4 & 89.6 & 83.7 & 81.1 \\
& + 50k Real Data & \textbf{96.5} & \textbf{94.4} & \textbf{96.3} & \textbf{95.6} & \textbf{86.2} & \textbf{92.4} & \textbf{87.2} & \textbf{84.7} \\
\bottomrule
\end{tabular}
}
\end{table*}

\textbf{Upper bound of GAN.} We are further interested in the upper bound of our method. As our method is designed based on adversarial training, the limitations of the GAN cause some failure cases. As illustrated in Figure \ref{fig:failure}, the well-trained generator fails to generate character patterns on difficult samples, particularly when the source image is of low quality and the curvature of the text shape is too high. One possible reason is the mode-dropping phenomenon studied by Bau et al.\  \citep{bau2019seeing}. Another reason is the lingering gap observed by \citep{Zhu2017hr} between the training supervision of paired and unpaired samples. To break this ceiling, one possible solution is to improve the synthesis engine and integrate various paired lifelike samples for training. This may lead to substantially more powerful generators, but heavily dependent on the development of synthesis engines. 

\begin{table*}[t]
\centering
\caption{Comparisons of generation in RGB space and in gray.}
\label{table:RGB}
\setlength{\tabcolsep}{5mm}{
\begin{tabular}{c c c c c c c c c}
\toprule
\multirow{2}{*}{Space} & \multicolumn{4}{c}{Regular Text} & \multicolumn{4}{c}{Irregular Text} \\
\cmidrule(lr){2-5} \cmidrule(lr){6-9}
& IIIT5K & SVT & IC03 & IC13 & SVT-P & CUTE & IC15-S & IC15 \\
\midrule
Gray & \textbf{95.4} & \textbf{92.7} & \textbf{96.3} & 94.8 & \textbf{85.4} & \textbf{89.6} & \textbf{83.7} & \textbf{81.1} \\
RGB & 95.2 & 92.3 & \textbf{96.3} & \textbf{95.6} & 85.3 & 88.9 & \textbf{83.7} & \textbf{81.1} \\
\bottomrule
\end{tabular}
}
\end{table*}

Inspired by recent work \citep{shi2018aster,Liao2019Mask}, it is possible to integrate several outputs of the system and choose the most possible one to achieve performance gain. Therefore, we proposed a simple yet effective method to address the issue stated above. The source image and the corresponding generated image are concatenated as a batch for network inference. Then, we choose the prediction with the higher confidence. As shown in the last row in Tables \ref{table:Results on general benchmarks} and \ref{table:Results on Irregular} (noted as ``+Com."), this ensemble mechanism greatly boosts the system performance, which indicates that the source and generated images are complementary to each other.

\subsection{More Accessible Data}
In the experiments of comparing the proposed method with previous recognition methods, we have used only synthetic data for fair comparison. Here, we use the ASTER \citep{shi2018aster} to explore whether there is room for improvement in synthesis engines.

Following Li et al.\  \citep{li2018show}, we collect publicly  available  real data for training. In contrast to synthetic data, real data is more costly to collect and annotate. Thus, there are only approximately 50k public real samples for training, whereas there are millions of synthetic data. As shown in Table \ref{table:real-data}, after we add the small real training set to the large synthetic one, the generality of both the baseline ASTER and our method is further boosted. This suggests that synthetic data is not sufficiently real and the model is still data-hungry. 

In summary, our approach is able to make full use of real samples in the wild to further gain robustness, because of the training of our method requires only input images and the corresponding text labels. Note that our method trained using only synthetic data even outperforms the baseline trained using real data on most benchmarks, particularly on SVT-P ($\uparrow$6.0\%). Therefore, noisy background style normalization is a promising way to improve recognition performance.

\subsection{Discussion}

\textbf{Generation in RGB space or in gray.} The background noise and text content may be relative easier to be separated in RGB colorful images. To this end, we conduct an experiment to evaluate the influence of RGB color space. The target style samples are synthesized in random color to guide the generation in RGB space. As shown in Table \ref{table:RGB}, we find that the generation in RGB space cannot outperform the generation in gray. Therefore, the key issue of background normalization is not the color space, but the lack of pixel-level supervision. Without fine-grained guidance at pixel level, the generation is only guided by the attention mechanism of the recognizer to focus informative regions. Other noisy regions on the generated image are unreasonably neglected.

\textbf{Alignment issue on long text.} To tackle the lack of paired training samples, we exploit the attention mechanism to extract every character for adversarial training. However, there exists misalignment problems of the attention mechanism \citep{cheng2017focusing,bai2018edit}, especially on long text. \citep{Cong2019Comparative} conducted a comprehensive study on the attention mechanism and found that the attention-based recognizers have poor performance on text sentence recognition. Thus, our method still have scope for performance gains on text sentence recognition. This is a common issue of most attention mechanisms, which merits further study.

\section{Conclusion}
\label{section:Conclusion}

We have presented a novel framework for scene text recognition from a brand new perspective of separating text content from noisy background styles. The proposed method can greatly reduce recognition difficulty and thus boost the performance dramatically. Benefiting from the interactive joint training of an attention-based recognizer and a generative adversarial architecture, we extract character-level features for further adversarial training. Thus the discriminator focuses on informative regions and provides effective guidance for the generator. Moreover, the discriminator learns from the confusion of the recognizer and further effectively guides the generator. Thus, the generated patterns are clearer and easier to read. This feedback mechanism contributes to the generality of the generator. Our framework is end-to-end trainable, requiring only the text images and corresponding labels. Because of the elegant design, our method can be flexibly integrated with recent mainstream recognizers to achieve new state-of-the-art performance.

The proposed method is a successful attempt to solve the scene text recognition problem from the brand new perspective of image generation and style normalization, which has not been addressed intensively before. In the future, we plan to extend the proposed method to deal with end-to-end scene text recognition. How to extend our method to multiple general object recognition is also a topic of interest.

\begin{acknowledgements}

This research was in part supported in part by NSFC (Grant No. 61771199, 61936003), GD-NSF (No. 2017\-A\-0\-3\-0\-3\-1\-2\-00\-6), the National Key Research and Development Program  of China (No. 2016YFB1001405), Guangdong Intellectual Property Office Project (2018-10-1), and Guangzhou Science, Technology and Innovation Project (201704020134).

\end{acknowledgements}

\small 
\bibliographystyle{spbasic}
\bibliography{mybibfile}

%
%


\end{document}